\documentclass[twoside,11pt]{article}

\usepackage{jmlr2e}
\usepackage{xcolor}
\usepackage{makecell}
\usepackage{booktabs}
\usepackage{tcolorbox}
\PassOptionsToPackage{hyphens}{url}\usepackage{hyperref}
\usepackage{lastpage}
\usepackage{listings}

\definecolor{dkred}{rgb}{0.5,0,0}
\definecolor{dkgreen}{rgb}{0,0.6,0}
\definecolor{gray}{rgb}{0.5,0.5,0.5}
\definecolor{mauve}{rgb}{0.58,0,0.82}

\jmlrheading{20}{2019}{1-\pageref{LastPage}}{1/19; Revised
4/19}{5/19}{19-011}{Yue Zhao, Zain Nasrullah and Zheng Li}

\ShortHeadings{PyOD: A Python Toolbox for Scalable Outlier Detection}{Zhao, Nasrullah and Li}

\firstpageno{1}

\begin{document}

\title{PyOD: A Python Toolbox for Scalable Outlier Detection}

\author{\name Yue Zhao \email zhaoy@cmu.edu \\
       \addr Carnegie Mellon University\thanks{Work initialized while at University of Toronto.}\\
       Pittsburgh, PA 15213, USA
       \AND  
       \name Zain Nasrullah \email znasrullah@cs.toronto.edu\\
       \addr University of Toronto\\
       Toronto, ON M5S 2E4, Canada
       \AND
       \name Zheng Li \email jk\_zhengli@hotmail.com\\
       \addr Northeastern University Toronto\\
       Toronto, ON M5X 1E2, Canada}

\editor{Alexandre Gramfort}

\maketitle

\begin{abstract}
\texttt{PyOD} is an open-source Python toolbox for performing scalable outlier detection on multivariate data. Uniquely, it provides access to a wide range of outlier detection algorithms, including established outlier ensembles and more recent neural network-based approaches, under a single, well-documented API designed for use by both practitioners and researchers. With robustness and scalability in mind, best practices such as unit testing, continuous integration, code coverage, maintainability checks, interactive examples and parallelization are emphasized as core components in the toolbox's development. \texttt{PyOD} is compatible with both Python 2 and 3 and can be installed through Python Package Index (PyPI) or \url{https://github.com/yzhao062/pyod}.

\end{abstract}


\section{Introduction}

Outlier detection, also known as anomaly detection, refers to the identification of rare items, events or observations which differ from the general distribution of a population. Since the ground truth is often absent in such tasks, dedicated outlier detection algorithms are extremely valuable in fields which process large amounts of unlabelled data and require a means to reliably perform pattern recognition \citep{akoglu2012fast}. Industry applications include fraud detection in finance \citep{Ahmed2016survey}, fault diagnosis in mechanics \citep{Shin2005one}, 
intrusion detection in network security \citep{Garcia2009anomaly} and pathology detection in medical imaging \citep{baur2018deep}.

To help approach these problems, established outlier detection packages exist in various programming languages such as \texttt{ELKI Data Mining} \citep{Achtert2010visual} and 
\texttt{RapidMiner} \citep{Hofmann2013rapidminer} in Java and \texttt{outliers} \citep{Komsta2011outliers} in R. However, Python, one of the most important languages in machine learning, still lacks a dedicated toolkit for outlier detection. Existing implementations either stand as single algorithm tools like \texttt{PyNomaly} \citep{Constantinou2018pynomaly} or exist as part of a general-purpose framework like \texttt{scikit-learn} \citep{Pedregosa2011scikit} which does not cater specifically to anomaly detection. To fill this gap, we propose and implement \texttt{PyOD}---a comprehensive 
Python toolbox for scalable outlier detection.

Compared to existing libraries, \texttt{PyOD} has six distinct advantages. Firstly, it
contains more than 20 algorithms which cover both classical techniques such as local outlier factor and recent neural network architectures such as autoencoders or adversarial models. Secondly, \texttt{PyOD} implements combination methods for merging the results of multiple detectors and outlier ensembles which are an emerging set of models. Thirdly, \texttt{PyOD} includes a unified API, detailed documentation and interactive examples across all algorithms for clarity and ease of use. Fourthly, all models are covered by unit testing with cross platform continuous integration, code coverage and code maintainability checks. Fifthly, optimization instruments are employed when possible: just-in-time (JIT) compilation and parallelization are enabled in select models for scalable outlier detection. Lastly, \texttt{PyOD} is compatible with both Python 2 and 3 across major operating systems (\texttt{Windows}, \texttt{Linux} and \texttt{MacOS}). Popular detection algorithms implemented in \texttt{PyOD} (version 0.7.0) are summarized in Table \ref{table:algorithms}. 

\begin{table}
\centering
	\begin{tabular}{lccc } 
	    \toprule
		\textbf{Method} &
		\textbf{Category} &
		\makecell{\textbf{JIT} \\ \textbf{Enabled}} &
		\makecell{\textbf{Multi-}   \\ \textbf{core}} \\
		\midrule

        LOF \citep{Breunig2000lof}                      & Proximity         & No    & Yes \\  
        kNN \citep{Ramaswamy2000efficient}              & Proximity         & No    & Yes \\
        AvgkNN \citep{Angiulli2002fast}                 & Proximity         & No    & Yes \\
        CBLOF \citep{He2003discovering}                 & Proximity         & Yes   & No  \\  
        OCSVM \citep{scholkopf2001estimating}           & Linear Model      & No    & No  \\ 
        LOCI \citep{Papadimitriou2003loci}              & Proximity         & Yes   & No  \\
        PCA \citep{Shyu2003novel}                       & Linear Model      & No    & No  \\          
        MCD \citep{Hardin2004outlier}                   & Linear Model      & No    & No  \\  
        Feature Bagging \citep{Lazarevic2005feature}	& Ensembling		& No	& Yes \\
        ABOD \citep{Kriegel2008angle}                   & Proximity         & Yes   & No  \\
        Isolation Forest \citep{Liu2008isolation}	    & Ensembling		& No	& Yes \\
        HBOS \citep{Goldstein2012histogram}             & Proximity         & Yes   & No  \\
        SOS \citep{janssens2012stochastic}              & Proximity         & Yes   & No  \\
        AutoEncoder \citep{Sakurada2014anomaly}         & Neural Net        & Yes   & No  \\
        AOM \citep{Aggarwal2015theoretical}             & Ensembling        & No    & No  \\
        MOA \citep{Aggarwal2015theoretical}             & Ensembling        & No    & No  \\
        SO-GAAL \citep{Liu2019generative}               & Neural Net        & No    & No  \\
        MO-GAAL \citep{Liu2019generative}               & Neural Net        & No    & No  \\
        XGBOD \citep{Zhao2018xgbod}    				    & Ensembling     	& No	& Yes \\
        LSCP \citep{Zhao2019lscp}				        & Ensembling		& No	& No  \\

		\bottomrule
	\end{tabular}
	\caption{Select outlier detection models in \texttt{PyOD}} 
	\label{table:algorithms} 
\end{table}

\section{Project Focus}

\textbf{Build robustness}. Continuous integration tools (\textit{Travis CI}, 
\textit{AppVeyor}  
and \textit{CircleCI}) 
are leveraged to conduct automated testing under various versions of Python and operating systems. 
Tests are executed daily, when commits are made to the development and master branches, or when a pull request is opened.\\ 
\textbf{Quality assurance}. The project follows the \texttt{PEP8} standard; maintainability is ensured by \textit{CodeClimate}, 
an automated code review and quality assurance tool. Additionally, code blocks with high cognitive complexity are actively refactored and a standard set of unit tests exist to ensure 95\% overall code coverage. These design choices enhance collaboration and this standard is enforced on all pull requests for consistency. \\
\textbf{Community-based development}. \texttt{PyOD}'s code repository is hosted on GitHub\footnote{\url{https://github.com/yzhao062/pyod}} to facilitate collaboration. At the time of this writing, eight contributors have helped develop the code base and others have contributed in the form of bug reports and feature requests.\\
\textbf{Documentation and examples}. Comprehensive documentation is developed using \texttt{sphinx} and \texttt{numpydoc} and rendered using \textit{Read the Docs}\footnote{\url{https://pyod.readthedocs.io}}. It includes detailed API references, an installation guide, code examples and algorithm benchmarks. An interactive Jupyter notebook is also hosted on \textit{Binder}
allowing others to experiment prior to installation. \\
\textbf{Project relevance}. \texttt{PyOD} has been used in various academic and commercial projects \citep{Zhao2018dcso,ramakrishnan2019anomaly,krishnan2019alphaclean}. The GitHub repository receives more than 10,000 monthly views and its PyPI downloads exceed 6,000 per month.

\section{Library Design and Implementation}
\texttt{PyOD} is compatible with both Python 2 and 3 using \texttt{six}; it relies on \texttt{numpy}, \texttt{scipy} and \texttt{scikit-learn} as well. Neural networks such as autoencoders and SO\_GAAL additionally require \texttt{Keras}. To enhance model scalability, select algorithms (Table \ref{table:algorithms}) are optimized with JIT using \texttt{numba}. Parallelization for multi-core execution is also available for a set of algorithms using \texttt{joblib}. Inspired by \texttt{scikit-learn}'s API design \citep{buitinck2013api}, all implemented outlier detection algorithms inherit from a base class with the same interface: (i) \texttt{fit} processes the train data and computes the necessary statistics; (ii) \texttt{decision\_function} generates raw outlier scores for unseen data after the model is fitted; (iii) \texttt{predict} returns a binary class label corresponding to each input sample instead of the raw outlier score and (iv) \texttt{predict\_proba} offers the result as a probability using either normalization or Unification \citep{Kriegel2011interpreting}. Within this framework, new models are easy to implement by taking advantage of inheritance and polymorphism. Base methods can then be overridden as necessary. 

Once a detector has been fitted on a training set, the corresponding train outlier scores and binary labels are accessible using its \texttt{decision\_scores\_} and \texttt{labels\_} attributes. Once fitted, the model's \texttt{predict}, \texttt{decision\_function} and \texttt{predict\_proba} methods may be called for use on new data. An example showcasing the ease of use of this API is shown in Code Snippet 1 with an angle-based outlier detector (ABOD).

\begin{lstlisting}[title={Code Snippet 1: Demo of PyOD API with the ABOD detector},captionpos=b]
  >>> from pyod.models.abod import ABOD
  >>> from pyod.utils.data import generate_data
  >>> from pyod.utils.data import evaluate_print
  >>> from pyod.utils.example import visualize
  >>>
  >>> X_train, y_train, X_test, y_test = generate_data(\
  ...     n_train=200, n_test=100, n_features=2)
  >>>
  >>> clf = ABOD(method="fast")                              # initialize detector
  >>> clf.fit(X_train)                               
  >>>
  >>> y_test_pred = clf.predict(X_test)                      # binary labels
  >>> y_test_scores = clf.decision_function(X_test)          # raw outlier scores
  >>> y_test_proba = clf.predict_proba(X_test)               # outlier probability
  >>>
  >>> evaluate_print("ABOD", y_test, y_test_scores)          # performance evaluation
  ABOD Performance; ROC: 0.934; Precision at n: 0.902
  >>>
  >>> visualize(y_test, y_test_scores)                       # prediction visualization 
  
\end{lstlisting}
\vskip 0.2in

\begin{figure*}[hb]
	\includegraphics[width=\linewidth]{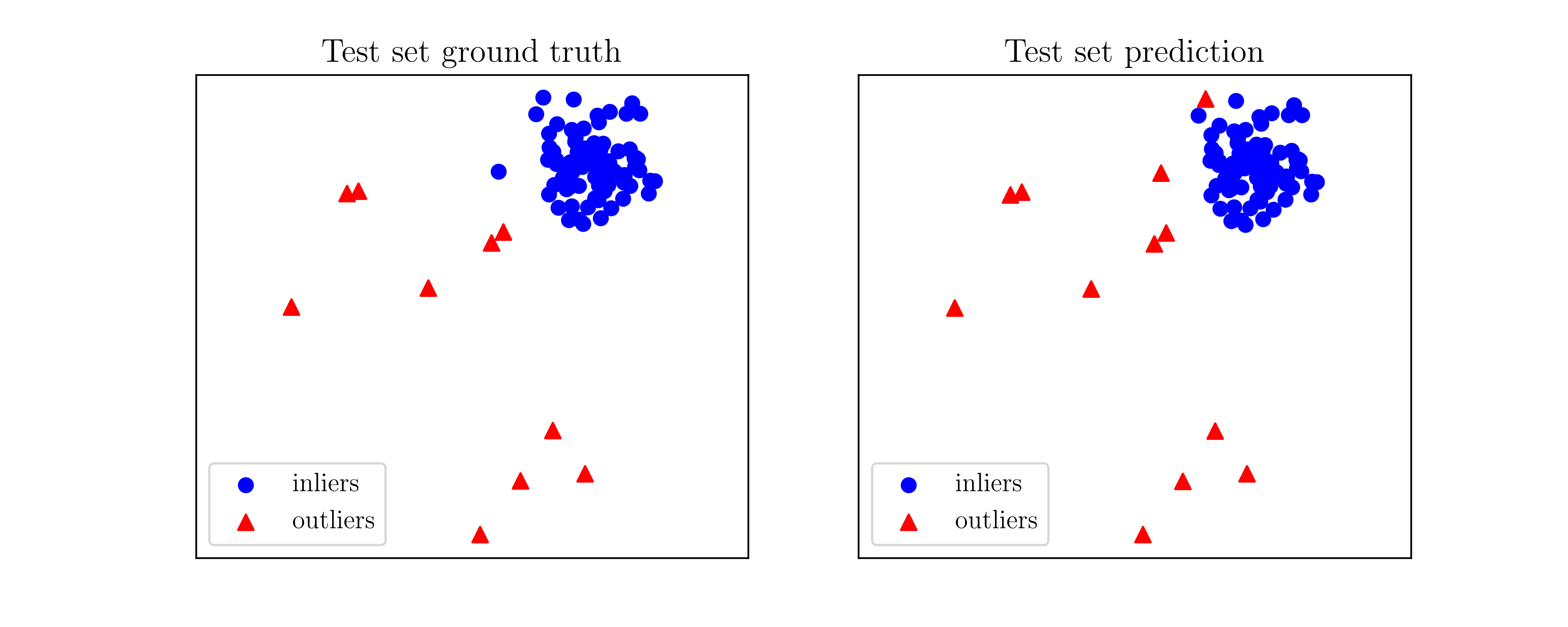}
	\caption{Demonstration of using \texttt{PyOD} in visualizing prediction result}
	\label{fig:vis}
\end{figure*}

In addition to the outlier detection algorithms, a set of helper and utility functions (\texttt{generate\_data}, \texttt{evaluate\_print} and \texttt{visualize}) are included in the library for quick model exploration and evaluation. The two-dimensional artificial data used in the example is created by \texttt{generate\_data} which generates inliers from a Gaussian distribution and outliers from a uniform distribution. An example of using \texttt{visualize} is shown in Figure \ref{fig:vis}.

\newpage

\section{Conclusion and Future Plans}
This paper presents \texttt{PyOD}, a comprehensive toolbox built in Python for scalable outlier detection. It includes more than 20 
classical and emerging detection algorithms and is being used in both academic and commercial projects. As avenues for future work, we
plan to enhance the toolbox by implementing models that work well with time series and geospatial data, improving computational efficiency through distributed computing and addressing engineering challenges such as handling sparse matrices or memory limitations.


\acks
We thank the editor and anonymous reviewers for their constructive comments. This work was partly supported by Mitacs through the Mitacs Accelerate program. 

\vskip 0.2in
\bibliography{ref}

\end{document}